\documentclass[11pt]{article}

\usepackage{acl}
\usepackage{times}
\usepackage{latexsym}
\usepackage{microtype}
\usepackage{graphicx}
\usepackage{amsmath}
\usepackage{amssymb}
\usepackage{booktabs}
\usepackage{array}
\usepackage{colortbl}
\usepackage{longtable}
\usepackage{enumitem}
\usepackage{placeins}
\usepackage{listings}
\usepackage{inconsolata}
\usepackage[T1]{fontenc}
\usepackage[utf8]{inputenc}

\title{\textsl{Fund2Persona}: A Framework for Building and Refining Financial Advisor Personas from Fund Disclosure Data}

\author{\textbf{Suhwan Park$^{1}$\footnotemark[1], Hoyoung Lee$^{1,2}$\footnotemark[1], Zhangyang Wang$^{3}$,} \\
\textbf{Alejandro Lopez-Lira$^{4}$, Young Cha$^{5}$, Chanyeol Choi$^{2}$, Jaewon Choi$^{6}$\footnotemark[2], Yongjae Lee$^{1,2}$\footnotemark[2]} \\[0.45em]
$^{1}$UNIST \quad $^{2}$LinqAlpha \quad $^{3}$University of Texas at Austin \\
$^{4}$University of Florida \quad $^{5}$Blackstone \quad $^{6}$Hanwha Life}

\makeatletter
\g@addto@macro\@thanks{%
  \footnotetext[1]{Equal contribution.}%
  \footnotetext[2]{\raggedright Corresponding authors: jaewonch@hanwha.com; yongjaelee@unist.ac.kr.}%
  \begingroup
  \footnotetext{\vspace{1pt}Earlier version: arXiv:2606.29793.}%
  \addtocounter{footnote}{-1}%
  \endgroup
}
\makeatother

\newcommand{\persona}{\theta}
\newcommand{\method}[1]{\textbf{#1}}
\newcommand{\sd}[1]{{\scriptsize$\pm$#1}}
\newcommand{\statcell}[2]{\shortstack{#1\\[-1pt]{\scriptsize #2}}}
\definecolor{quantheader}{RGB}{222,235,247}
\definecolor{qualheader}{RGB}{226,239,218}
\definecolor{methodgroup}{RGB}{235,235,235}
\definecolor{lightrow}{RGB}{247,247,247}
\definecolor{promptbg}{RGB}{250,247,241}
\newcommand{\methodgrouprow}[1]{\rowcolor{methodgroup}\multicolumn{7}{@{}l@{}}{\textbf{\textit{#1}}}}
\newcommand{\filterstep}[1]{\textbf{#1}}
\newcommand{\filterlast}[1]{\textbf{#1}}
\lstdefinestyle{promptstyle}{
  basicstyle=\ttfamily\scriptsize,
  breaklines=true,
  columns=fullflexible,
  frame=single,
  framerule=0.25pt,
  rulecolor=\color{methodgroup},
  backgroundcolor=\color{promptbg},
  xleftmargin=3pt,
  xrightmargin=3pt,
  aboveskip=4pt,
  belowskip=5pt,
  showstringspaces=false,
  keepspaces=true
}

\AtBeginDocument{%
  \setlength{\abovedisplayskip}{6pt plus 2pt minus 2pt}%
  \setlength{\belowdisplayskip}{6pt plus 2pt minus 2pt}%
  \setlength{\abovedisplayshortskip}{6pt plus 2pt minus 2pt}%
  \setlength{\belowdisplayshortskip}{6pt plus 2pt minus 2pt}%
}

\begin{document}
\maketitle

\begin{abstract}
Demand for personalized financial advice is growing, yet current LLM-based advisors often fail to provide consistent and specialized guidance.\ Simple persona prompts rarely specify how a financial advisor should reason and often drift toward generic recommendations. We propose \textbf{Fund2Persona}, a framework that builds financial-advisor personas from real-world fund disclosures and refines them through an actor--scorer--patcher loop. We test whether the resulting personas can predict held-out portfolio changes and produce commentary consistent with fund managers' own explanations. They outperform generic baselines on both tasks. We further study two downstream diagnostics: market-scenario generation, where persona retrieval broadens plausible investment views, and multi-turn investor--advisor conversations, where matched personas give more specific and useful advice than a generic advisor. These results suggest that real fund data can bring manager-specific investment expertise to LLM advisors rather than merely changing an LLM's surface style.
\end{abstract}

\section{Introduction}

Large language models (LLMs) are increasingly used not only as general-purpose assistants but also as specialized systems designed for particular roles and areas of expertise \citep{kim2023analysts}. This is especially appealing for financial advising, where users want guidance that reflects coherent views about risk, horizon, and investment style. However, access to consistent specialized expertise remains limited. A common way to emulate such expertise is persona prompting, where a short label or system prompt assigns the model an identity and the model is expected to answer in character. Yet such persona conditioning is often under-specified: personas placed in system prompts do not reliably improve task behavior \citep{zheng2024helpful}, and automatically generated personas often fail to reflect the intended traits and preferences \citep{li2025promise}.

This problem is particularly important in finance, where LLMs are increasingly used as financial advisors and investing agents \citep{ross2024economicus,ross2025onesize,li2026finsaber}. For users, an advisor's investment perspective directly shapes how they weigh risk, time horizon, and conviction when making personalized portfolio recommendations. Users need advice that reflects coherent investment logic rather than generic heuristics, yet LLM financial advisors often collapse to average recommendations driven by a dominant factor \citep{ross2025onesize}. This creates a need for LLM financial advisors that reflect a specific manager's investment perspective and decision-making style rather than relying on a generic role label alone.

\begin{figure*}[t]
  \centering
  \includegraphics[width=0.98\textwidth]{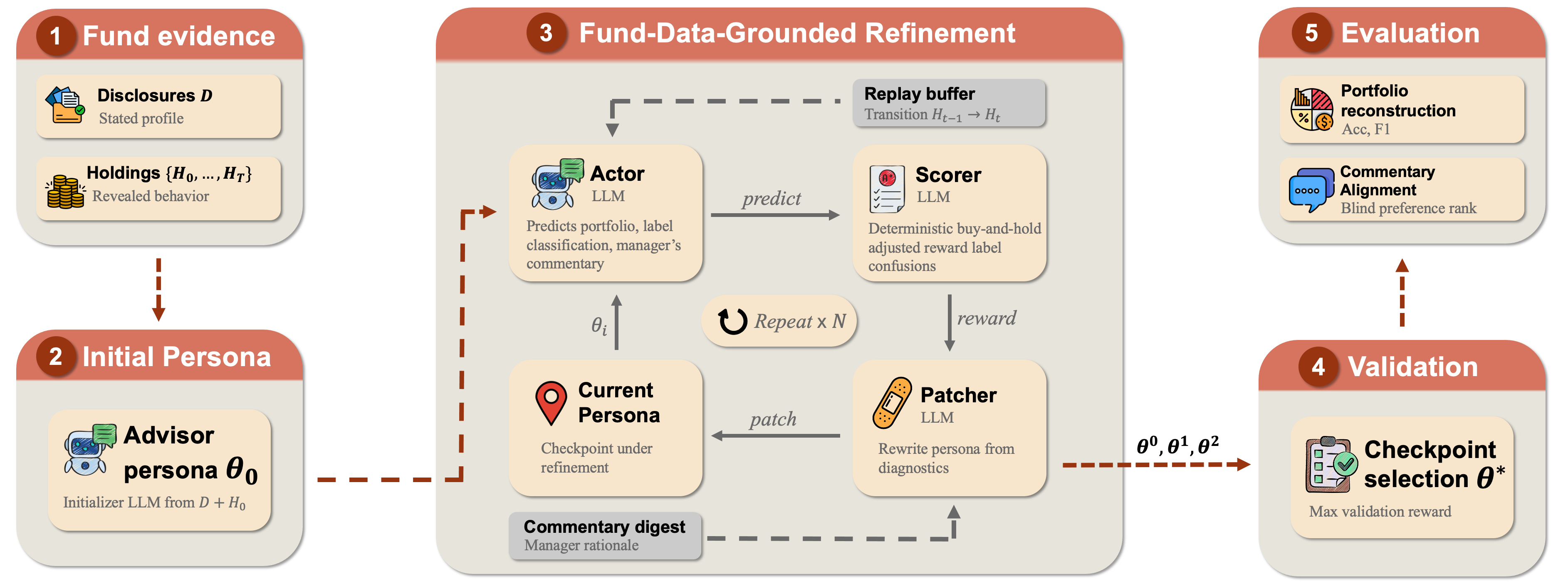}
  \caption{Framework overview. Fund2Persona builds an initial persona from fund evidence, performs fund-data-grounded refinement with replay transitions and manager commentary, selects a checkpoint by validation, and evaluates it on a held-out period.}
  \label{fig:framework}
\end{figure*}

We argue that a fund is a natural source for such grounding: its disclosure filings state the manager's investment philosophy, while repeated holdings snapshots reveal how that philosophy translates into portfolio decisions. We propose \textbf{Fund2Persona}, a framework that synthesizes a financial-advisor persona from fund disclosure data and holdings snapshots and refines it with past holdings transitions, market context, and manager commentary. Figure~\ref{fig:framework} summarizes the overall pipeline.

In summary, our contributions are as follows.
\begin{itemize}[leftmargin=*,itemsep=2pt,topsep=2pt]
  \item We propose \textbf{Fund2Persona}, a framework that uses fund disclosures to build a reusable financial-advisor persona that captures a manager's investment perspective and decision-making style.
  \item We design a fund-data-grounded refinement and evaluation protocol that tests whether the persona can reconstruct held-out portfolio changes and produce commentary consistent with the actual fund manager's explanations.
  \item We also test the personas in market-scenario generation and multi-turn financial advice, where they broaden market perspectives and provide more specific guidance than a generic advisor.
\end{itemize}

\FloatBarrier

\section{Related Work}

\paragraph{\mbox{Behavioral Persona Modeling.}}

Prompting can readily shape LLM behavior for specific tasks and domains, but meaningful persona construction still requires more than assigning a short role label. The key challenge is grounding the persona in stable behavioral evidence. Underspecified personas do not reliably produce distinct task behavior \citep{zheng2024helpful}, while LLM-generated personas can fail to capture the multi-dimensional attributes they are intended to represent \citep{li2025promise,wang2024flatten}. To give personas more concrete traits and histories, prior work builds them from interviews with real people \citep{park_generative_2024}, demographic profiles and backstories \citep{argyle_out_2023,moon_virtual_2024}, or large synthetic persona datasets \citep{castricato_persona_2025}. These approaches broaden persona construction, but often require costly elicitation or indirect grounding. We instead refine personas from fund data through repeated feedback and revision, following prior work on self-refinement and reflective language agents \citep{madaan_self-refine_2023,shinn_reflexion_2023}.

\paragraph{Financial-Advisor Personas.}

Financial advice depends on more than factual knowledge; investors also care about how an advisor understands their goals and communicates trade-offs. Investors often rely on trusted advisors \citep{gennaioli2015money}, and studies of generative-AI advisors show that preference elicitation, investment style, and advisor personality affect both suitability and trust \citep{takayanagi2025personalized}. Recent work has explored LLMs for financial decision making, investment advice, and investing agents \citep{ross2024economicus,ross2025onesize,ross2025breaking,li2026finsaber}. However, their recommendations can be prompt-sensitive or overly generic. Existing work focuses largely on investor personalization, without explicitly modeling the advisor's underlying investment logic. Fund2Persona addresses this gap by grounding each advisor persona in evidence from a real fund, giving it a distinct investment approach rather than a loosely specified role.

\section{Framework}

Our framework converts fund disclosures and holdings snapshots into a financial-advisor persona that reflects the manager's investment decision policy, refines it with holdings-transition feedback and manager commentary, and evaluates the fixed persona on a held-out period. The whole process is summarized in Figure~\ref{fig:framework}.

\subsection{Problem Setup: Fund as a Policy Object}

A fund $\mathcal{F}$ provides disclosure text $D$ and quarterly holdings snapshots $\{H_t\}_{t=0}^{T}$. We use these sources to construct a financial-advisor persona $\persona$. To prevent leakage, we use replay transitions $H_{t-1}\!\rightarrow\!H_t$ to refine the persona, then select the best checkpoint on a separate validation period. The persona is held fixed for evaluation. For each transition, candidates are limited to the starting snapshot's top-10 holdings. We compare the following methods.
\begin{itemize}[leftmargin=*,topsep=2pt,itemsep=0pt,parsep=0pt,partopsep=0pt]
  \item \method{Buy-and-Hold}: carries the observed holdings forward without active trading, so portfolio weights change only with realized returns.
  \item \method{Generic LLM}: makes portfolio decisions from the task context alone, without a fund persona.
  \item \method{Disclosure-Only Persona}: constructs the persona from the fund disclosure alone, without historical holdings or refinement feedback.
  \item \method{Initial Fund Persona} ($\persona_0$): combines the fund disclosure and initial holdings into a persona before transition-based refinement.
  \item \method{Random-Fund Persona}: applies a refined persona randomly sampled from a different fund to the target fund's portfolio.
  \item \method{Fund2Persona (ours)} ($\persona^*$): refines the initial persona using historical portfolio changes and selects the best checkpoint on validation data.
\end{itemize}

\begin{table*}[!t]
\centering
\small
\renewcommand{\arraystretch}{1.08}
\setlength{\tabcolsep}{2.5pt}
\begin{tabular*}{\textwidth}{@{\extracolsep{\fill}}lcccccc@{}}
\toprule
& \multicolumn{4}{c}{\cellcolor{quantheader}\textbf{Quantitative}}
& \multicolumn{2}{c}{\cellcolor{qualheader}\textbf{Qualitative}} \\
& \multicolumn{4}{c}{\cellcolor{quantheader}\textit{Portfolio reconstruction}}
& \multicolumn{2}{c}{\cellcolor{qualheader}\textit{Commentary alignment}} \\
\cmidrule(lr){2-5}\cmidrule(lr){6-7}
Method & 3-Acc $\uparrow$ & 3-F1 $\uparrow$ & 5-Acc $\uparrow$ & 5-F1 $\uparrow$ & rank-1 (\%) $\uparrow$ & avg rank $\downarrow$ \\
\midrule
\methodgrouprow{No-persona baselines} \\
Buy-and-Hold    & \underline{0.416}\sd{0.000} & 0.196\sd{0.000} & 0.303\sd{0.000} & 0.093\sd{0.000} & -- & -- \\
Generic LLM     & 0.338\sd{0.008} & 0.334\sd{0.009} & 0.278\sd{0.012} & 0.204\sd{0.021} & 17.5 & 3.625 \\
\midrule
\methodgrouprow{Persona-conditioned methods} \\
Disclosure-Only Persona & 0.352\sd{0.019} & 0.349\sd{0.019} & 0.282\sd{0.005} & 0.221\sd{0.009} & 15.0 & 3.150 \\
Initial Fund Persona    & 0.367\sd{0.019} & 0.362\sd{0.019} & 0.292\sd{0.009} & 0.239\sd{0.011} & \underline{20.0} & \underline{2.600} \\
Random-Fund Persona     & 0.393\sd{0.014} & \underline{0.364}\sd{0.002} & \underline{0.308}\sd{0.008} & \underline{0.243}\sd{0.003} & 17.5 & 3.175 \\
\textbf{Fund2Persona (ours)} & \textbf{0.441\sd{0.007}} & \textbf{0.410\sd{0.010}} & \textbf{0.346\sd{0.004}} & \textbf{0.270\sd{0.003}} & \textbf{30.0} & \textbf{2.450} \\
\bottomrule
\end{tabular*}
\caption{Held-out fund-faithfulness results. Quantitative entries report mean $\pm$ standard deviation over three inference runs. Random-Fund Persona uses an independently sampled persona from another fund in each repeat. Portfolio reconstruction evaluates 69 funds over the top-10 holdings observed at $H_6$ and scores the rolled-forward portfolios at the $H_8$ endpoint; commentary alignment reports results for the 40 funds with available manager-commentary data. Full confidence intervals and paired tests are reported in Appendix~\ref{app:repeated-uncertainty}.}
\label{tab:results}
\end{table*}

\subsection{Initial Persona Construction}

An initializer LLM builds $\persona_0$ from disclosure $D$ and initial holdings $H_0$. The persona follows a structure covering the fund's investment beliefs, portfolio construction, trading style, responses to market events, and risk posture. At this point, it is based only on the disclosure and initial portfolio.

\subsection{Holdings-Transition Reconstruction and Active-Delta Labels}

We use the same persona-conditioned actor for refinement and evaluation. For each transition $H_{t-1}\!\rightarrow\!H_t$, it receives the persona, the starting portfolio, candidate holdings, realized fund and stock returns, and market context. It then predicts each candidate's end-of-period portfolio weight and active-delta label, together with a brief rationale. The task is to recover the manager's portfolio actions, not to forecast asset returns.

An increase in portfolio weight does not necessarily mean that the manager increased exposure: the weight may have risen simply because the asset outperformed. To separate price movement from active positioning, we first compute the weight that each holding would have reached with no trading:
\begin{equation}
w^{\mathrm{BH}}_i = B_0\,\tfrac{w^{0}_i(1+R_i)}{\sum_{j\in\mathcal{C}} w^{0}_j(1+R_j)},
\end{equation}
where $R_i$ is the holding's return, $\mathcal{C}$ is the fixed top-10 candidate universe at the starting snapshot, and $B_0=\sum_{\scriptscriptstyle j\in\mathcal{C}}w_j^0$ preserves its total weight in the fund portfolio. This lets us decompose the observed weight change into a price-driven component and a manager-action component:
\begin{equation}
\begin{aligned}
\Delta^{\mathrm{price}}_i &= w^{\mathrm{BH}}_i - w^0_i, \\
\Delta^{\mathrm{active}}_i &= w^{\mathrm{gt}}_i - w^{\mathrm{BH}}_i, \\
w^{\mathrm{gt}}_i - w^0_i &= \Delta^{\mathrm{price}}_i + \Delta^{\mathrm{active}}_i .
\end{aligned}
\end{equation}
The active delta is the behavioral object we score: it is positive when the reported end weight is above price drift, negative when it is below price drift, and near zero when the manager largely lets price drift operate. The actor's prediction is judged on the same buy-and-hold-adjusted scale, $\widehat{\Delta}^{\mathrm{active}}_i=\hat{w}_i-w^{\mathrm{BH}}_i$. We assign one of five behavioral labels: \method{active increase} when the manager raises exposure beyond price drift, \method{drift-following} when the position moves with returns, \method{target-maintaining} when trading offsets drift to preserve the starting weight, \method{active decrease} when the manager reduces exposure, and \method{exit} when the position is removed. Since the candidate universe is the start-snapshot top-10, entry is not a reported label in this evaluation. On this buy-and-hold-adjusted scale, a 0.25 percentage-point threshold effectively filters negligible residual changes. We also report a 3-class version: \method{active increase}, \method{maintain}, and \method{active decrease}. Detailed class distributions and full confusion matrices are reported in Appendix~\ref{app:active-delta-diagnostics}.

\subsection{Reward: Buy-and-Hold-Adjusted Improvement}

Let $d(\cdot,\cdot)$ denote the portfolio distance between two weight vectors. Because buy-and-hold already captures price-driven weight changes, the reward measures how much the actor improves on this baseline:
\begin{equation}
r \;=\; \frac{d(w^{\mathrm{BH}},w^{\mathrm{gt}}) - d(\hat{w},w^{\mathrm{gt}})}{d(w^{\mathrm{BH}},w^{\mathrm{gt}})}.
\end{equation}
$r>0$ means the actor is closer to the fund's reported end holdings than buy-and-hold. For refinement, we combine this reward with active-delta label errors and the actor's rationales to identify why a decision was wrong.

\subsection{Data-Grounded Patch Update}

A deterministic scorer summarizes replay errors using the buy-and-hold-adjusted reward, active-delta labels, and predicted-versus-actual top-10 weights. The patcher updates the persona from this feedback together with the actor's rationale and the actual manager's commentary. Two rounds produce the checkpoint sequence $\persona_0 \rightarrow \persona_1 \rightarrow \persona_2$.

\subsection{Selecting the Final Persona}

We compare $\persona_0$, $\persona_1$, and $\persona_2$ on a separate validation period rather than selecting a checkpoint from replay performance. The actor evaluates all three checkpoints on the same validation transitions, and we retain the checkpoint with the highest average buy-and-hold-adjusted reward. The selected persona is kept unchanged for held-out evaluation.

\section{Experiments}
\subsection{Experimental Setup}

\noindent\textbf{Data.} To keep comparisons consistent, we retain 69 funds with complete disclosure, holdings, and aligned price data over the same evaluation periods. \textbf{SEC Form 497K prospectuses} describe each fund's mandate, risk posture, and investment universe. \textbf{N-PORT filings} provide nine quarterly portfolio snapshots, which define the initial portfolio, replay and validation transitions, held-out holdings, and each top-10 candidate set. \textbf{Daily price and return data} for each fund and its holdings are used to calculate buy-and-hold drift, active deltas, rewards, and tracking error. Each month, the actor also receives a \textbf{summary of relevant macroeconomic, sector, and company events}. Finally, \textbf{N-CSR/N-CSRS shareholder reports} provide the managers' actual commentary for persona refinement and qualitative evaluation. \hyperref[app:fund-selection]{Appendix~A} details the data sources and filtering process.

\vspace{3pt}
\noindent\textbf{Pipeline and Model.} We evaluate our framework in two held-out settings: a quantitative task that measures how well it reconstructs portfolio decisions, and a qualitative task that measures how closely its output matches the actual manager's market interpretation. The initial persona is built from $H_0$; replay uses transitions through $H_4$, validation uses the next two transitions ending at $H_6$, and held-out evaluation rolls the portfolio forward from $H_6$ to $H_8$. \texttt{GPT-5.4 Mini} is used for persona construction, iterative patching, and judging; \texttt{Gemini 3.1 Flash-Lite} is used for replay, validation, held-out actor evaluation, and dialogue/scenario generation. Held-out evaluation occurs entirely after the actor model's knowledge cutoff.

\label{sec:metrics}
\noindent\textbf{Metrics.} The quantitative evaluation uses active-delta label metrics reused from \S3.3--3.4. We report \textit{Accuracy} ($\uparrow$), the fraction of correctly predicted labels for $K\in\{3,5\}$ classes, $\mathrm{Acc}_K=\frac{1}{N}\sum_i \mathbf{1}[\hat{y}_i^{(K)}=y_i^{(K)}]$, and \textit{Macro-F1} ($\uparrow$), the mean per-class F1, which is robust to label imbalance and is likewise reported for both class settings.

\begin{figure*}[t]
  \centering
  \includegraphics[width=0.98\textwidth]{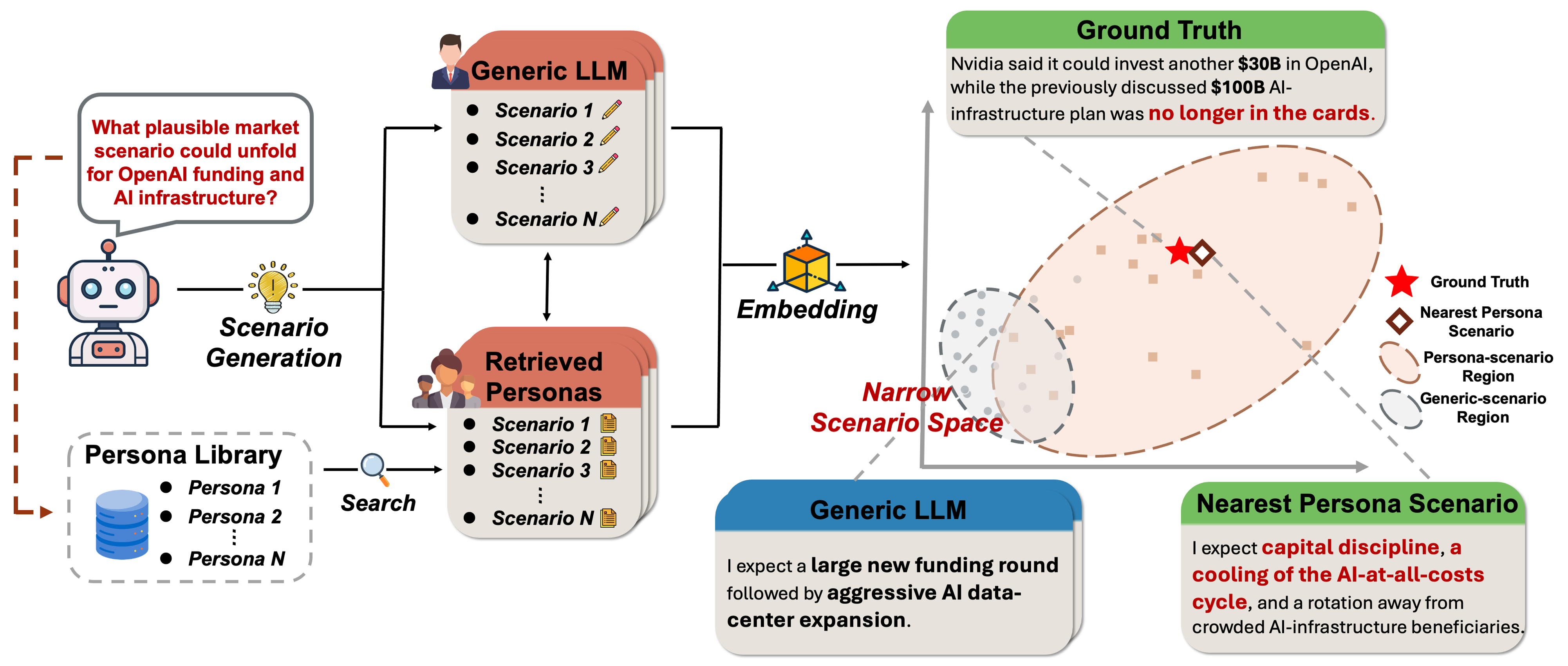}
  \caption{Scenario-space coverage for market forecasting. For the same open-ended market query, generic LLM rollouts concentrate around a narrow funding-and-buildout thesis, while agentically retrieved fund personas generate a broader set of persona-conditioned scenarios. The resolved outcome lies near a persona scenario that anticipates a slowdown in OpenAI's infrastructure spending plans.}
  \label{fig:scenario-coverage}
\end{figure*}

\subsection{Quantitative: Post-Period Copy-Trading Reconstruction}

\noindent\textbf{Task.} We test whether each method can reproduce a fund's later portfolio decisions over the 2025Q3--2026Q1 held-out interval. Starting from the fund's observed top-10 holdings, each method updates a simulated portfolio using the same monthly market context and realized fund and holding returns. We then directly compare the final simulated weights with the fund's reported holdings.

\vspace{3pt}
\noindent\textbf{Results.} Table~\ref{tab:results} reports the held-out reconstruction metrics averaged over repeated inference runs. Our approach shows the strongest overall reconstruction performance across the evaluated methods. Full paired tests are reported in Appendix~\ref{app:repeated-uncertainty}.

\subsection{Qualitative: Manager-Commentary Alignment}

\noindent\textbf{Task.} Within the held-out interval, we evaluate the 40 funds with available N-CSR/N-CSRS shareholder commentary for 2025Q3--2025Q4. Each method's persona generates manager-style commentary from the same market and portfolio context, which is compared directly against actual manager commentary. We evaluate manager-specific investment lens, risk posture, and market interpretation, not surface style. Detailed prompts are provided in Appendix~\ref{app:commentary-generation-prompt}--\ref{app:commentary-alignment-prompt}.

\vspace{3pt}
\noindent\textbf{Results.} A blind judge ranks the five methods' anonymized memos against the actual manager commentary by closeness to the manager's view. We report each method's rank-1 rate ($\uparrow$) and average rank ($\downarrow$). Because memos and commentary differ in information, this measures alignment with the manager's perspective, not whether the model used the manager's decision process. Table~\ref{tab:results} shows that Fund2Persona (ours) achieves the highest rank-1 rate (30.0\%) and best average rank (2.450).

\subsection{Refinement Ablations}
\label{sec:refinement-ablations}

We ablate refinement depth on reconstruction and shareholder-commentary feedback on manager-commentary alignment. We evaluate every saved checkpoint on the same held-out reconstruction task. $\persona^*$ is the checkpoint chosen on the separate validation period. To measure the effect of shareholder commentary, we keep the initial persona and replay feedback unchanged and remove only the commentary from the patcher's input. Full protocol details are provided in Appendix~\ref{app:refinement-ablation}.

\begin{table}[!h]
\centering
\footnotesize
\renewcommand{\arraystretch}{1.05}
\setlength{\tabcolsep}{1.5pt}
\begin{tabular*}{\columnwidth}{@{\extracolsep{\fill}}lcccc@{}}
\toprule
Variant & 3-Acc $\uparrow$ & 3-F1 $\uparrow$ & 5-Acc $\uparrow$ & 5-F1 $\uparrow$ \\
\midrule
$\persona_0$ & 0.367 & 0.362 & 0.292 & 0.239 \\
$\persona_1$ & 0.417 & 0.371 & 0.320 & 0.246 \\
$\persona_2$ & 0.422 & 0.380 & 0.333 & 0.259 \\
$\persona^*$ & \textbf{0.441} & \textbf{0.410} & \textbf{0.346} & \textbf{0.270} \\
\bottomrule
\end{tabular*}
\caption{Checkpoint ablation. $\persona^*$ is selected from $\persona_0$, $\persona_1$, and $\persona_2$ on the validation period.}
\label{tab:refinement-ablation}
\end{table}

The validation-selected persona performs best overall on reconstruction, suggesting that separate validation helps identify the best refinement checkpoint. We separately ablate the use of the manager's historical commentary during refinement on the qualitative alignment task. In blind comparisons against held-out manager commentary, the version refined with this feedback is preferred in 59.6\% of judgments, versus 40.4\% without it, suggesting that manager commentary helps the persona capture the manager's investment perspective.

\FloatBarrier

\section{Case Study}

\subsection{Persona-Guided Scenario Diversification}

\noindent\textbf{Setup.} The first case study evaluates scenario generation through a diversity lens: if each persona captures a distinct investment view, asking different personas the same question should produce a wider range of plausible scenarios. We use OpenForesight, a news-grounded benchmark in which models forecast future events from information before the outcomes are known \citep{chandak2025scaling}. We filter the dataset to 95 economic and business-related market-event queries; for each query, the Generic LLM baseline answers 20 times without a persona, while \method{Fund2Persona} retrieves 20 relevant fund personas and each retrieved persona generates one scenario. We embed the generated scenarios with \texttt{all-MiniLM-L6-v2} and measure scenario-space spread by the mean pairwise cosine distance within each 20-scenario set.

\begin{table}[!h]
\centering
\scriptsize
\setlength{\tabcolsep}{2pt}
\begin{tabular*}{\linewidth}{@{\extracolsep{\fill}}lccc@{}}
\toprule
Metric & Generic & Personas & Win \\
\midrule
Pairwise dist. $\uparrow$ &
0.313$\pm$0.070 &
\textbf{0.381$\pm$0.038} &
91.6\% \\
\bottomrule
\end{tabular*}
\caption{Scenario diversity across 95 OpenForesight market-event queries, measured as the mean pairwise cosine distance between scenario embeddings. Both methods generate 20 scenarios per query.}
\label{tab:scenario-diversity}
\end{table}

\noindent\textbf{Results.} Table~\ref{tab:scenario-diversity} shows that persona-conditioned generation expands the scenario space on aggregate. Under the same 20-scenario budget, retrieved personas increase the mean pairwise cosine distance from $0.313$ to $0.381$ and outperform the generic baseline on $91.6\%$ of the 95 queries. Although some increase in diversity is expected when different personas are used, this broader coverage has practical value by making a wider range of financial perspectives available for the same query. Figure~\ref{fig:scenario-coverage} visualizes the corresponding scenario embeddings after UMAP projection and provides a qualitative example of this aggregate pattern.

\subsection{Personalized Financial Advice}

\begin{figure}[!t]
  \centering
  \includegraphics[width=\linewidth]{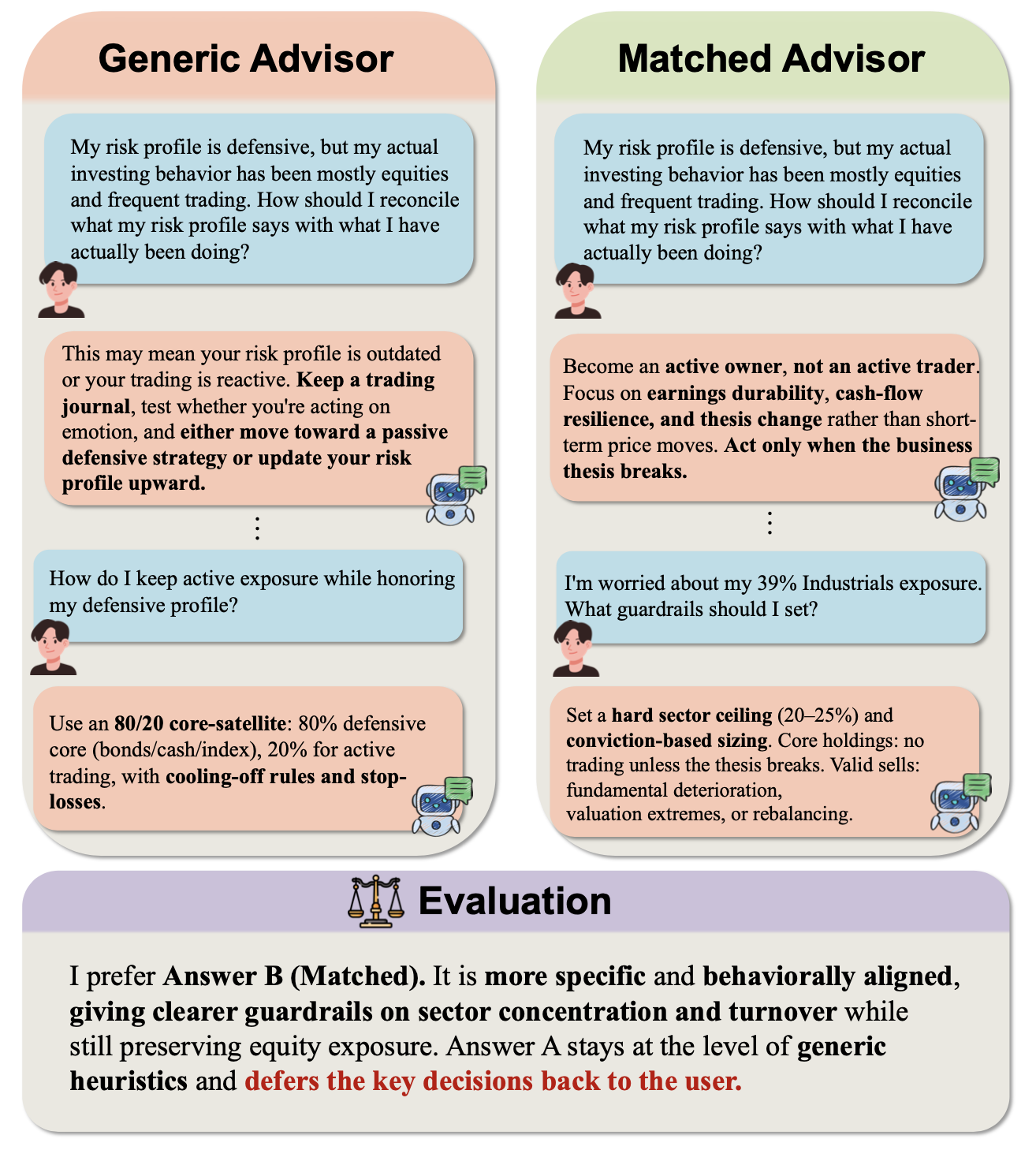}
  \caption{Example multi-turn advisory dialogue comparing a generic financial advisor with a matched fund-data-grounded financial advisor for the same investor profile. The bottom panel shows the blinded judge's pairwise judgment result.}
  \label{fig:advisory-dialogue}
\end{figure}

\noindent\textbf{Setup.} Individual investors differ in risk attitudes, constraints, holdings, and preferences, and they seek advice that translates those personal conditions into specialized investment guidance. We therefore ask whether matching investors with our advisor personas leads to more personalized and specialized financial advice. To evaluate personalized advice in a multi-turn setting, we need realistic investor profiles rather than synthetic preference lists. We use FAR-Trans customer attributes, transaction histories, and asset metadata to synthesize each investor profile and a plausible initial investment query \citep{sanzcruzado2024fartrans}.

From profiles that express an investment preference or tension, we randomly sample 50 investors. For each investor, an LLM-based search uses the profile and initial query to retrieve the advisor persona whose investment logic best matches the investor's needs. We compare this matched advisor with a generic advisor and a deliberately mismatched persona whose risk posture or investment preferences conflict with the investor's needs.

The user simulator, initialized with the same profile and preference summary, conducts five-turn dialogues with matched, generic, and mismatched advisors using \texttt{Gemini 3.1 Flash-Lite}. We also compare Fund2Persona with matched Disclosure-Only and Initial personas to separate refinement from fund matching; \texttt{GPT-5.4 Mini} judges all pairwise comparisons. All advisory prompts are provided in Appendix~\ref{app:advisory-persona-prompt}--\ref{app:advisory-judge-prompts}.

\begin{table}[h]
\centering
\footnotesize
\renewcommand{\arraystretch}{1.08}
\setlength{\tabcolsep}{4pt}
\setlength{\aboverulesep}{0.35ex}
\setlength{\belowrulesep}{0.35ex}
\begin{tabular*}{\columnwidth}{@{\extracolsep{\fill}}lrr@{}}
\toprule
Pairwise comparison & Turn 3 $\uparrow$ & Turn 5 $\uparrow$ \\
\midrule
Fund2Persona vs. Generic & 51.0 & 69.0 \\
Fund2Persona vs. Mismatched & 82.0 & 87.0 \\
Fund2Persona vs. Disclosure-Only & 72.0 & 65.0 \\
Fund2Persona vs. Initial Persona & 72.0 & 70.0 \\
\bottomrule
\end{tabular*}
\caption{Pairwise win rates of Fund2Persona for 50 investor profiles. The judge compares responses at turns 3 and 5 using both A/B presentation orders.}
\label{tab:case}
\end{table}

\noindent\textbf{Results.} Table~\ref{tab:case} shows that Fund2Persona exceeds 50\% in all four pairwise comparisons. Its advantage over the Generic advisor suggests that persona conditioning produces more concrete and personalized guidance than generic advice. Because the matched Disclosure-Only and Initial controls preserve fund identity, these comparisons isolate the contribution of persona refinement. Fund2Persona is consistently preferred to both controls at turns 3 and 5, indicating that the advisory gains are not explained by fund matching alone.

\subsection{Human Validation of LLM Judges}
\label{sec:human-eval}

We conduct a pilot human evaluation to check whether the LLM judges make the same choices as people. The participants include five finance-industry professionals and ten people with prior investment or fund experience, each of whom evaluates 10 randomly sampled instances per task. For commentary ranking, we measure whether the human and LLM judges select the same top-ranked response; for the personalized financial advice task, we measure whether they prefer the same response in each pair. Agreement is above chance in both tasks, providing initial evidence that the LLM judges capture the same overall preference direction as human evaluators. Full protocol details are provided in Appendix~\ref{app:human-eval}.

\begin{table}[h]
\centering
\footnotesize
\renewcommand{\arraystretch}{1.08}
\setlength{\tabcolsep}{4pt}
\begin{tabular}{@{}lcc@{}}
\toprule
Evaluation & Agreement & Chance \\
\midrule
Commentary ranking & \textbf{52.7\%} & 20.0\% \\
Personalized advice & \textbf{78.7\%} & 50.0\% \\
\bottomrule
\end{tabular}
\caption{Human--LLM agreement: top-1 agreement for commentary ranking and pairwise agreement for personalized financial advice. Each task includes 15 evaluators and 10 randomly assigned instances per evaluator.}
\label{tab:human-agreement}
\end{table}

\FloatBarrier

\section{Conclusion}

We introduced \textbf{Fund2Persona}, a framework that builds financial-advisor personas from real-world fund data. Rather than assigning an LLM a generic advisor role, it grounds each persona in the fund's investment perspective and decisions. An actor--scorer--patcher loop refines the persona by comparing its decisions with manager behavior while separating passive price drift from active changes. In out-of-sample evaluations, Fund2Persona better reproduced managers' portfolio decisions and market interpretations. It also broadened the range of plausible scenarios and provided more personalized advice when paired with investors. Together, these results show that real fund behavior can ground LLM advisors in distinct investment logic that remains useful across portfolio analysis, scenario generation, and personalized advice.

\section*{Limitations}

Our evaluation has two main limitations. First, the framework uses specific LLMs: GPT-5.4 Mini for persona construction, patching, and judging, and Gemini 3.1 Flash-Lite for replay, validation, held-out evaluation, and dialogue/scenario generation. Fund2Persona is model-agnostic, so practitioners can substitute models for their use case, but results may vary because LLMs can exhibit different tendencies in financial analysis \citep{lee2025yourai,kong2026bias}. Second, strict data requirements limit our study to 69 active U.S. equity funds. We require aligned holdings, disclosures, prices, and manager evidence so that every method is evaluated with the same information over the same periods. This makes the comparison more controlled, but narrows the available sample. We exclude passive funds because their portfolios largely follow predefined rules rather than discretionary manager decisions. We leave fixed-income, multi-asset, and non-U.S. funds to future work because comparable longitudinal data were not consistently available.

\section*{Ethical Considerations}

Fund2Persona is evaluated as a research prototype through simulated advisory dialogues and has not yet been deployed to real investors or evaluated by licensed financial advisors. As with other LLM-based financial systems, its outputs may be incomplete or inaccurate. Practical use therefore requires appropriate caution and qualified human oversight.

The human evaluation included five unpaid finance-industry professionals recruited through the authors' professional networks and ten Prolific participants with prior investment experience; the latter received approximately \$21 (about \$23 per hour). All provided informed consent before participation; we retained only anonymized ratings and response times, with no directly identifying or sensitive personal data. Formal IRB review was not required by the participating institutions.

\section*{Acknowledgments}

This work was supported by the National Research Foundation of Korea (NRF) grant funded by the Korea government (MSIT) (No.~RS-2025-24803208) and the Institute of Information \& Communications Technology Planning \& Evaluation (IITP) grant funded by the Korea government (MSIT) (No.~RS-2020-II201336, Artificial Intelligence Graduate School Program (UNIST)).

\FloatBarrier
\clearpage

\bibliography{reference}

\clearpage
\appendix
\onecolumn
\raggedbottom

\section{Data Construction}
\label{app:fund-selection}

\subsection{Fund Universe}
\label{app:fund-universe}

We begin with SEC fund series and apply the filters in Table~\ref{tab:fund-filter}. The aim is not to cover every U.S. fund, but to keep funds with the filings, holdings history, and price data needed to run the same experiment from start to finish.

\begin{table}[h]
\centering
\scriptsize
\setlength{\tabcolsep}{3pt}
\renewcommand{\arraystretch}{1.2}
\begin{tabular}{@{}>{\raggedright\arraybackslash}p{0.21\textwidth}>{\raggedright\arraybackslash}p{0.69\textwidth}>{\raggedleft\arraybackslash}p{0.065\textwidth}@{}}
\toprule
\rowcolor{methodgroup}
\textbf{Screen} & \textbf{Rationale} & \textbf{Funds} \\
\midrule
\filterstep{SEC fund series} & SEC series identifiers provide a common key for linking filings, disclosures, and tickers. & 11{,}858 \\
\filterstep{Nine N-PORT snapshots} & Keep funds with all nine quarterly snapshots from 2024Q1 through 2026Q1. & 9{,}932 \\
\filterstep{ETF-like public ticker} & Keep standard 2--4 letter tickers and remove mostly five-letter mutual-fund share classes, which are not directly comparable to the traded active and thematic funds studied here. & 2{,}345 \\
\filterstep{Trackable U.S. equity fund} & Require at least 80\% equity, at least 80\% U.S. exposure, and less than 5\% derivatives so that portfolio changes are visible mainly through stock holdings. & 936 \\
\filterstep{Usable 497K disclosure} & Require enough prospectus text to identify the fund's mandate and risk posture. & 908 \\
\filterstep{LLM mandate filter} & Retain active or thematic equity strategies and remove index trackers, fixed-rule products, and non-equity vehicles. & 118 \\
\filterlast{Complete price coverage} & Require prices for each fund and its top-10 holdings throughout the evaluation period. & 69 \\
\bottomrule
\end{tabular}
\caption{Fund selection process. The filters leave 69 active U.S. equity funds with complete data for the full experimental period.}
\label{tab:fund-filter}
\end{table}

The resulting 69 funds each have nine holdings snapshots, a disclosure excerpt, prices for the fund and its top holdings, monthly market summaries, and shareholder-report commentary where available. Using this common set lets us run every method on the same dates and inputs.

\begin{table}[!h]
\centering
\begingroup
\fontsize{7.2pt}{8.1pt}\selectfont
\setlength{\tabcolsep}{2pt}
\renewcommand{\arraystretch}{0.98}
\begin{tabular}{@{}>{\bfseries\raggedright\arraybackslash}p{0.055\textwidth}>{\raggedright\arraybackslash}p{0.43\textwidth}>{\bfseries\raggedright\arraybackslash}p{0.055\textwidth}>{\raggedright\arraybackslash}p{0.43\textwidth}@{}}
\toprule
\rowcolor{methodgroup}
\textbf{Ticker} & \textbf{Fund name} & \textbf{Ticker} & \textbf{Fund name} \\
\midrule
ACGR & American Century Sustainable Growth ETF & ACLC & American Century Sustainable Equity ETF \\
ACVF & American Conservative Values ETF & AMID & Argent Mid Cap ETF \\
AMOM & QRAFT AI-Enhanced U.S. Large Cap Momentum ETF & AOTG & AOT Growth and Innovation ETF \\
ARKQ & ARK Autonomous Technology \& Robotics ETF & AVIE & Avantis Inflation Focused Equity ETF \\
AVLV & Avantis U.S. Large Cap Value ETF & AVSU & Avantis Responsible U.S. Equity ETF \\
AVUS & Avantis U.S. Equity ETF & BCHP & Principal Focused Blue Chip ETF \\
BDVG & iMGP Berkshire Dividend Growth ETF & BEEZ & Honeytree U.S. Equity ETF \\
BUYZ & Franklin Disruptive Commerce ETF & DIVL & Madison Dividend Value ETF \\
DSTL & Distillate U.S. Fundamental Stability \& Value ETF & DUSA & Davis Select U.S. Equity ETF \\
DVND & Touchstone ETF Trust-Touchstone Dividend Select ETF & EMLP & First Trust North American Energy Infrastructure Fund \\
FBCG & Fidelity Blue Chip Growth ETF & FBCV & Fidelity Blue Chip Value ETF \\
FDG & American Century Focused Dynamic Growth ETF & FDV & Federated Hermes U.S. Strategic Dividend ETF \\
FFLC & Fidelity New Millennium ETF & FFLG & Fidelity Growth Opportunities ETF \\
FLV & American Century Focused Large Cap Value ETF & FMAG & Fidelity Magellan ETF \\
FPWR & First Trust EIP Carbon Impact ETF & GK & AdvisorShares Gerber Kawasaki ETF \\
HFGO & Hartford Large Cap Growth ETF & ITAN & Sparkline Intangible Value ETF \\
IWFG & IQ Winslow Focused Large Cap Growth ETF & IWLG & IQ Winslow Large Cap Growth ETF \\
JHAC & John Hancock Fundamental All Cap Core ETF & JTEK & JPMorgan U.S. Tech Leaders ETF \\
LCF & Trust-Touchstone US Large Cap Focused ETF & LGRO & Level Four Large Cap Growth Active ETF \\
LRGC & AB US Large Cap Strategic Equities ETF & LSGR & Natixis Loomis Sayles Focused Growth ETF \\
MDLV & Morgan Dempsey Large Cap Value ETF & MID & American Century Mid Cap Growth Impact ETF \\
MMLG & First Trust Multi-Manager Large Growth ETF & NDVG & Nuveen Dividend Growth ETF \\
NUGO & Nuveen Growth Opportunities ETF & NWLG & Nuveen Winslow Large-Cap Growth ESG ETF \\
OALC & ONEASCENT LARGE CAP CORE ETF & PJFG & PGIM Jennison Focused Growth ETF \\
PJFV & PGIM Jennison Focused Value ETF & QLTY & GMO U.S. Quality ETF \\
QRFT & QRAFT AI-Enhanced U.S. Large Cap ETF & RFFC & ALPS Active Equity Opportunity ETF \\
SEMI & Columbia Seligman Semiconductor and Technology ETF & SMRI & Bushido Capital US Equity ETF \\
STNC & Hennessy Stance ESG ETF & TBG & TBG Dividend Focus ETF \\
TCAF & T. Rowe Price Capital Appreciation Equity ETF & TCHP & T. Rowe Price Blue Chip Growth ETF \\
TEQI & T. Rowe Price Equity Income ETF & TGLR & LAFFER TENGLER Equity Income ETF \\
TGRT & T. Rowe Price Growth ETF & TGRW & T. Rowe Price Growth Stock ETF \\
TSME & Thrivent Small-Mid Cap ESG ETF & VNSE & Natixis Vaughan Nelson Select ETF \\
VSLU & Applied Finance Valuation Large Cap ETF & WUGI & AXS Esoterica NextG Economy ETF \\
XDAT & Franklin Exponential Data ETF & YALL & God Bless America ETF \\
ZECP & Zacks Earnings Consistent Portfolio ETF & & \\
\bottomrule
\end{tabular}
\endgroup
\caption{Funds included in the study.}
\label{tab:ticker-list}
\end{table}

\clearpage
\subsection{Temporal Coverage and Splits}

The nine snapshots cover 2024Q1--2026Q1. We use $H_0$ through $H_4$ for replay and $H_4$ through $H_6$ to select the final checkpoint. Held-out evaluation runs from $H_6$ to $H_8$.
\begin{center}
\small
\setlength{\tabcolsep}{4pt}
\renewcommand{\arraystretch}{1.15}
\begin{tabular}{@{}>{\bfseries\raggedright\arraybackslash}p{0.19\textwidth}>{\raggedright\arraybackslash}p{0.43\textwidth}>{\raggedright\arraybackslash}p{0.30\textwidth}@{}}
\toprule
\rowcolor{methodgroup}
Block & Holdings horizon & Use \\
\midrule
Replay & $H_0\rightarrow H_4$ (2024Q1--2025Q1) & Persona refinement feedback \\
Validation & $H_4\rightarrow H_6$ (2025Q1--2025Q3) & Checkpoint selection \\
Held-out evaluation & $H_6\rightarrow H_8$ (2025Q3--2026Q1) & Evaluation at final endpoint \\
\bottomrule
\end{tabular}
\end{center}

\subsection{Disclosure, Quarterly Holdings, and Prices}

Fund profiles come from 497K prospectuses downloaded from SEC EDGAR and converted from filing HTML to plain text. N-PORT filings are linked by SEC series identifier, with the reported top-10 equity positions retained from each quarterly snapshot. Daily prices are then matched to the fund ticker and each of these holdings.

\subsection{Market and News Context}

A monthly market summary is constructed from public news articles. It covers the main macroeconomic, policy, sector, and company events for that month and is supplied during replay, validation, and held-out evaluation. The summaries exclude shareholder commentary, future holdings, labels, and scorer feedback.

\subsection{Shareholder Commentary Digests}

Annual N-CSR and semiannual N-CSRS shareholder reports are obtained from SEC EDGAR. Passages in which the manager explains market conditions, performance, or portfolio changes are retained, while administrative, holdings-only, and boilerplate text is removed. Commentary follows the same temporal split as the holdings data: the patcher receives only reports that were filed, and whose reporting periods ended, before validation began. Later reports are reserved as references for the qualitative evaluation, and no commentary is shown to the evaluation actor.

\subsection{Case-Study Diagnostic Queries}

The scenario-diversity study draws economic, business, and market questions from the OpenForesight HuggingFace release \citep{chandak2025scaling}. The source question and metadata remain fixed across the generic and persona-conditioned generators.

\begin{table}[!h]
\centering
\footnotesize
\setlength{\tabcolsep}{4pt}
\renewcommand{\arraystretch}{1.12}
\begin{tabular}{@{}>{\bfseries\raggedright\arraybackslash}p{0.10\textwidth}>{\raggedright\arraybackslash}p{0.84\textwidth}@{}}
\toprule
\rowcolor{methodgroup}
QID & Raw source question \\
\midrule
155813 & Which group will be cited as expecting U.S. soybean farmers to have a third straight year of losses in 2025, by December 31, 2025? \\
35 & Which partner group will face a 15\% baseline U.S. tariff on autos, pharmaceuticals, and semiconductors? \\
733 & Which nuclear power plant will receive GBP 14.2 billion in funding under the 2025 Spending Review? \\
\bottomrule
\end{tabular}
\caption{Examples of raw OpenForesight records used to build the scenario-diversity query set.}
\label{tab:openforesight-query-examples}
\end{table}

\FloatBarrier

\section{Experiment Details}
\label{app:experiment-details}

\subsection{Model Usage}

\begin{center}
\refstepcounter{table}\label{tab:model-usage}
\footnotesize
\setlength{\tabcolsep}{5pt}
\renewcommand{\arraystretch}{1.12}
\begin{tabular}{@{}p{0.23\linewidth}p{0.50\linewidth}p{0.18\linewidth}@{}}
\toprule
Model & Used for & Knowledge cutoff \\
\midrule
GPT-5.4 Mini & Persona construction, patching, and judging & Aug. 2025 \\
Gemini 3.1 Flash-Lite & Replay, validation, held-out actor evaluation, commentary, dialogue, and scenario generation & Jan. 2025 \\
all-MiniLM-L6-v2 & Scenario embeddings & N/A \\
\bottomrule
\end{tabular}
\par\vspace{3pt}
\small Table~\thetable: Model roles in the experiments. Knowledge cutoffs are provider-reported for the LLMs.
\end{center}

\subsection{Human Evaluation}
\label{app:human-eval}

The pilot study includes five finance-industry professionals and ten participants with prior investment or fund experience. For each qualitative task, every evaluator receives 10 instances sampled from the corresponding original evaluation pool. The interface anonymizes method identities and response order. Commentary agreement is exact top-1 agreement between the evaluator's ranking and the LLM ranking; advisory-dialogue agreement is exact agreement between their pairwise preferences. The interface does not permit ties, and only completed responses are included. Agreement is pooled over the completed assigned instances. Table~\ref{tab:human-agreement} reports the resulting human--LLM agreement rates.

\subsection{Refinement Ablations}
\label{app:refinement-ablation}

For the checkpoint ablation, we directly evaluate each saved checkpoint $\{\persona_0,\persona_1,\persona_2\}$ and the validation-selected $\persona^*$ three times on the same held-out reconstruction subset. For the commentary ablation, we keep the initial persona and replay feedback fixed and change only whether shareholder-commentary digests are available to the patcher. We evaluate the resulting personas under the same held-out reconstruction protocol. For the qualitative comparison, we use the 40 funds with held-out manager commentary, generate three commentaries per condition, and judge each pair four times with balanced A/B presentation order. The results are reported in Section~\ref{sec:refinement-ablations} and Table~\ref{tab:refinement-ablation}.

\subsection{Repeated-Inference Uncertainty}
\label{app:repeated-uncertainty}

We perform three inference runs for each generative condition. For \method{Random-Fund Persona}, each repeat independently assigns a final persona from another fund. Predictions are first summarized as a confusion matrix for each fund in each run. The fund-level matrices are then averaged across the three runs, and aggregate Accuracy and Macro-F1 are computed from the averaged counts. We obtain 95\% confidence intervals from 100{,}000 percentile-bootstrap samples that retain each fund and its top-10 holdings as a cluster. Paired $p$-values use 200{,}000 two-sided fund-cluster permutations against \method{Fund2Persona}. The complete run identifiers and Random-Fund assignment manifests are retained with the evaluation artifacts.

\begin{table}[!ht]
\centering
\footnotesize
\setlength{\tabcolsep}{2pt}
\renewcommand{\arraystretch}{1.14}
\begin{tabular*}{\textwidth}{@{\extracolsep{\fill}}lcccc@{}}
\toprule
Method & \shortstack{\textbf{3-Acc}\\[-1pt]{\scriptsize [95\% CI] ($p$)}} & \shortstack{\textbf{3-F1}\\[-1pt]{\scriptsize [95\% CI] ($p$)}} & \shortstack{\textbf{5-Acc}\\[-1pt]{\scriptsize [95\% CI] ($p$)}} & \shortstack{\textbf{5-F1}\\[-1pt]{\scriptsize [95\% CI] ($p$)}} \\
\midrule
Buy-and-Hold & \statcell{0.416 [0.368, 0.464]}{$p=0.2547$} & \statcell{0.196 [0.179, 0.211]}{\textbf{$p<0.0001^{***}$}} & \statcell{0.303 [0.257, 0.351]}{$p=0.0956$} & \statcell{0.093 [0.082, 0.104]}{\textbf{$p<0.0001^{***}$}} \\
\rowcolor{lightrow}
Generic LLM & \statcell{0.338 [0.310, 0.367]}{\textbf{$p=0.0003^{***}$}} & \statcell{0.334 [0.305, 0.361]}{\textbf{$p=0.0050^{**}$}} & \statcell{0.278 [0.249, 0.308]}{\textbf{$p=0.0105^{*}$}} & \statcell{0.204 [0.184, 0.224]}{\textbf{$p=0.0025^{**}$}} \\
Disclosure-Only Persona & \statcell{0.352 [0.326, 0.379]}{\textbf{$p=0.0005^{***}$}} & \statcell{0.349 [0.322, 0.376]}{\textbf{$p=0.0127^{*}$}} & \statcell{0.282 [0.256, 0.309]}{\textbf{$p=0.0055^{**}$}} & \statcell{0.221 [0.197, 0.245]}{\textbf{$p=0.0227^{*}$}} \\
\rowcolor{lightrow}
Initial Fund Persona & \statcell{0.367 [0.339, 0.395]}{\textbf{$p=0.0014^{**}$}} & \statcell{0.362 [0.332, 0.388]}{\textbf{$p=0.0215^{*}$}} & \statcell{0.292 [0.265, 0.319]}{\textbf{$p=0.0125^{*}$}} & \statcell{0.239 [0.211, 0.265]}{$p=0.1310$} \\
Random-Fund Persona & \statcell{0.393 [0.361, 0.425]}{\textbf{$p=0.0050^{**}$}} & \statcell{0.364 [0.337, 0.388]}{\textbf{$p=0.0043^{**}$}} & \statcell{0.308 [0.281, 0.334]}{\textbf{$p=0.0191^{*}$}} & \statcell{0.243 [0.222, 0.262]}{$p=0.0503$} \\
\textbf{Fund2Persona} & \statcell{\textbf{0.441 [0.407, 0.474]}}{--} & \statcell{\textbf{0.410 [0.380, 0.438]}}{--} & \statcell{\textbf{0.346 [0.314, 0.377]}}{--} & \statcell{\textbf{0.270 [0.244, 0.294]}}{--} \\
\bottomrule
\end{tabular*}
\caption{Primary repeated-inference uncertainty analysis. Values are means with fund-clustered 95\% confidence intervals; $p$ is the raw paired fund-level value against Fund2Persona. $^{*}p<0.05$, $^{**}p<0.01$, and $^{***}p<0.001$.}
\label{tab:repeated-uncertainty}
\end{table}

Fund2Persona has the highest mean performance across all four metrics, although some paired differences do not reach statistical significance. Buy-and-Hold is included as a deterministic no-trade reference rather than a persona-based method, so the main comparison concerns the learned and persona-conditioned baselines. We therefore interpret the results as an overall performance advantage rather than universal statistical superiority.

\FloatBarrier

\subsection{Fund-Specificity Controls}

Active U.S. equity funds share market context and broad active-management priors, which can make a randomly assigned fund persona informative even when it does not match the target manager. To test target-policy compatibility more directly, we compare the matched persona with Random-Fund Persona and with deliberately mismatched personas. For each target fund, three personas with contrasting disclosed objectives and management characteristics are applied under the same reconstruction protocol and their results are averaged.

\begin{table}[!ht]
\centering
\footnotesize
\setlength{\tabcolsep}{5pt}
\renewcommand{\arraystretch}{1.12}
\begin{tabular*}{0.88\textwidth}{@{\extracolsep{\fill}}lcccc@{}}
\toprule
Persona condition & 3-Acc & 3-F1 & 5-Acc & 5-F1 \\
\midrule
Matched Persona & \textbf{0.441} & \textbf{0.410} & \textbf{0.346} & \textbf{0.270} \\
Random-Fund Persona & 0.393 & 0.364 & 0.308 & 0.243 \\
Mismatched Personas & 0.369 & 0.350 & 0.291 & 0.234 \\
\bottomrule
\end{tabular*}
\caption{Reconstruction under matched, randomly assigned, and deliberately mismatched persona conditions.}
\label{tab:fund-specificity-controls}
\end{table}

Performance increases from deliberately mismatched personas to Random-Fund Persona and then to the matched persona. This pattern suggests that active funds share some common investment behavior, but a persona matched to the target fund reconstructs its decisions more accurately.

\FloatBarrier

\subsection{Active-Delta Class Diagnostics}
\label{app:active-delta-diagnostics}

Tables~\ref{tab:confusion-three-class} and~\ref{tab:confusion-five-class} report the Fund2Persona confusion matrices underlying the repeated-inference result. Each cell is the mean count across three inference runs, followed by its percentage within the actual-label row. The Total column gives the number and overall share of each actual class.

\begin{table}[!ht]
\centering
\footnotesize
\setlength{\tabcolsep}{5pt}
\renewcommand{\arraystretch}{1.12}
\begin{tabular*}{0.92\textwidth}{@{\extracolsep{\fill}}lrrrr@{}}
\toprule
Actual / Predicted & Active increase & Hold & Active decrease & Total \\
\midrule
Active increase & 48.7 (31.0\%) & 69.7 (44.4\%) & 38.7 (24.6\%) & 157 (22.8\%) \\
Hold & 63.0 (22.0\%) & 177.3 (61.8\%) & 46.7 (16.3\%) & 287 (41.6\%) \\
Active decrease & 62.3 (25.3\%) & 105.7 (43.0\%) & 78.0 (31.7\%) & 246 (35.7\%) \\
\bottomrule
\end{tabular*}
\caption{Three-class active-delta confusion matrix.}
\label{tab:confusion-three-class}
\end{table}

\begin{table}[!ht]
\centering
\scriptsize
\setlength{\tabcolsep}{3.5pt}
\renewcommand{\arraystretch}{1.12}
\begin{tabular*}{\textwidth}{@{\extracolsep{\fill}}lrrrrrr@{}}
\toprule
Actual / Predicted & Increase & Drift-following & Target-maintaining & Decrease & Exit & Total \\
\midrule
Active increase & 48.7 (31.0\%) & 51.3 (32.7\%) & 18.3 (11.7\%) & 38.7 (24.6\%) & 0.0 (0.0\%) & 157 (22.8\%) \\
Drift-following & 51.7 (24.7\%) & 93.3 (44.7\%) & 32.3 (15.5\%) & 31.7 (15.2\%) & 0.0 (0.0\%) & 209 (30.3\%) \\
Target-maintaining & 11.3 (14.5\%) & 26.7 (34.2\%) & 25.0 (32.1\%) & 15.0 (19.2\%) & 0.0 (0.0\%) & 78 (11.3\%) \\
Active decrease & 50.3 (23.3\%) & 59.0 (27.3\%) & 34.7 (16.0\%) & 71.7 (33.2\%) & 0.3 (0.2\%) & 216 (31.3\%) \\
Exit & 12.0 (40.0\%) & 8.0 (26.7\%) & 4.0 (13.3\%) & 6.0 (20.0\%) & 0.0 (0.0\%) & 30 (4.3\%) \\
\bottomrule
\end{tabular*}
\caption{Five-class active-delta confusion matrix. Column labels abbreviate active increase and active decrease where needed for space.}
\label{tab:confusion-five-class}
\end{table}

\FloatBarrier

The three-class labels comprise 22.8\% active increase, 41.6\% hold, and 35.7\% active decrease. In the five-class setting, drift-following and active decrease are the two largest classes, while exit is rare. The matrices therefore expose both the finer-grained drift/target distinction and the remaining class imbalance.

\subsection{Construction Cost}

Fund2Persona uses a fixed construction budget rather than refining until convergence. One initializer call produces $\persona_0$; two refinement rounds use four actor calls and one patcher call each; and validation evaluates three checkpoints on two transitions. End-to-end persona construction and validation selection therefore require 17 LLM calls per fund.

\begin{table}[!ht]
\centering
\footnotesize
\setlength{\tabcolsep}{8pt}
\renewcommand{\arraystretch}{1.12}
\begin{tabular}{@{}lr@{}}
\toprule
\rowcolor{methodgroup}
Computational item & Per fund \\
\midrule
Initializer & 1 call \\
Two refinement rounds & 10 calls \\
Validation selection & 6 calls \\
\textbf{Total construction and selection} & \textbf{17 calls} \\
Mean construction and selection tokens & 144.9K input + output tokens \\
Mean recorded provider cost & USD 0.078 \\
\bottomrule
\end{tabular}
\caption{Measured persona-construction and validation-selection cost per fund.}
\label{tab:construction-cost}
\end{table}

\FloatBarrier

\subsection{Top-10 Candidate-Universe Scope}

The evaluated funds contain 125 reported holdings on average (median 51). Applying the roll-forward actor to every reported position would enlarge the portfolio, market, and return context at every monthly step: the observed top-10 input averages approximately 4.7K tokens per call, whereas applying the same template to 125 positions is estimated at approximately 21.5K tokens, or 4.6 times as much input. We therefore use the starting top-10 as a fixed major-position universe. As a complementary diagnostic within this restricted universe, we compare the daily returns of the normalized top-10 portfolio with the realized fund return path. The normalized top-10 proxy yields 7.96\% annualized tracking error, indicating that the major-position proxy retains meaningful information about fund-level return dynamics.

\FloatBarrier

\subsection{Buy-and-Hold Drift and Active-Delta Examples}

For each transition, buy-and-hold first converts the starting weight $w_i^0$ into the return-drifted weight $w_i^{\mathrm{BH}}$. The price-driven change and the inferred behavior change are
\[
\Delta_i^{\mathrm{price}}=w_i^{\mathrm{BH}}-w_i^0,\qquad
\Delta_i^{\mathrm{active}}=w_i^{\mathrm{gt}}-w_i^{\mathrm{BH}}.
\]
The portfolio reward compares model error to the buy-and-hold error:
\[
D_{\mathrm{BH}}=\frac{1}{2}\sum_i |w_i^{\mathrm{gt}}-w_i^{\mathrm{BH}}|,\quad
D_{\mathrm{model}}=\frac{1}{2}\sum_i |w_i^{\mathrm{gt}}-\hat{w}_i|,\quad
r=\frac{D_{\mathrm{BH}}-D_{\mathrm{model}}}{D_{\mathrm{BH}}}.
\]
Table~\ref{tab:active-delta-examples} shows the same arithmetic on three actual held-out positions. Here $w^0$ is the starting weight, $w^{\mathrm{BH}}$ is the return-drifted buy-and-hold weight, $w^{\mathrm{gt}}$ is the next reported weight, and $\hat{w}$ is the model prediction; Price, Active, and Pred. are the corresponding deltas relative to $w^0$ or $w^{\mathrm{BH}}$. The final column, $r_i$, is the position-level analogue of the reward term; the portfolio reward $r$ uses the active-share sums above.

\begin{table}[!h]
\centering
\footnotesize
\setlength{\tabcolsep}{2pt}
\renewcommand{\arraystretch}{1.15}
\begin{tabular}{@{}>{\raggedright\arraybackslash}p{0.26\textwidth}*{8}{>{\centering\arraybackslash}p{0.075\textwidth}}@{}}
\toprule
\rowcolor{methodgroup}
Case & $w^0$ & $w^{\mathrm{BH}}$ & $w^{\mathrm{gt}}$ & $\hat{w}$ & Price & Active & Pred. & $r_i$ \\
\midrule
Active increase: GK--LLY & 3.68 & 4.68 & 7.19 & 7.29 & +0.99 & +2.52 & +2.61 & 0.96 \\
Target-maintaining: BEEZ--CMI & 4.12 & 5.95 & 4.28 & 4.28 & +1.83 & -1.67 & -1.67 & 1.00 \\
Active decrease: FLV--JNJ & 5.55 & 6.77 & 3.46 & 4.00 & +1.22 & -3.31 & -2.77 & 0.84 \\
\bottomrule
\end{tabular}
\caption{Examples of price drift, active delta, and buy-and-hold-adjusted reward arithmetic. Case names use fund ticker--holding ticker notation, e.g., GK--LLY is LLY within fund GK. Weights and deltas are percentage points of fund weight. Price $=w^{\mathrm{BH}}-w^0$; Active $=w^{\mathrm{gt}}-w^{\mathrm{BH}}$; Pred. $=\hat{w}-w^{\mathrm{BH}}$.}
\label{tab:active-delta-examples}
\end{table}

\FloatBarrier

\subsection{Generated Persona Examples}

\begin{center}
\refstepcounter{table}\label{tab:generated-persona-examples}
\scriptsize
\setlength{\tabcolsep}{3pt}
\renewcommand{\arraystretch}{1.13}
\begin{tabular}{@{}>{\raggedright\arraybackslash}p{0.08\textwidth}>{\raggedright\arraybackslash}p{0.18\textwidth}>{\raggedright\arraybackslash}p{0.70\textwidth}@{}}
\toprule
\rowcolor{methodgroup}
Persona & Style & Generated persona excerpt (shortened) \\
\midrule
\rowcolor{lightrow}\texttt{ARKQ} & Robotics and automation growth & I target long-term capital appreciation in companies leading automation, AI, energy transformation, and advanced materials. My approach is conviction-driven but disciplined: in market stress, policy uncertainty, or liquidity shocks, I trim high-beta or overvalued names more proactively to manage concentration and tracking error. In accommodative or pro-innovation regimes, I am more willing to maintain or add to high-conviction names, but I still avoid chasing positions that have already drifted far above their intended role. \\
\texttt{SEMI} & Semiconductor stack and AI infrastructure & I focus on the semiconductor stack and adjacent technology enablers, favoring product leadership, strong execution, and durable franchises across compute, chip design, manufacturing, test, and packaging. In strong AI-led or semiconductor-led tapes, I can add where the manager is intentionally leaning into leadership, but I still distinguish a true active increase from ordinary drift. When leadership becomes crowded, rates stay restrictive, or a winner is overextended, I prefer disciplined maintenance or trimming rather than reflexive winner-chasing. \\
\rowcolor{lightrow}\texttt{EMLP} & Energy infrastructure income & I specialize in North American energy infrastructure, emphasizing fee-for-service assets, pipelines, utilities, storage, and stable cash flows while limiting commodity-price cyclicality. I react to regime shifts only when they affect cash-flow stability or financing conditions: rising-rate or sticky-inflation backdrops push me toward pricing power, conservative balance sheets, and long-term contracts. In a falling-rate environment, I look for companies that could benefit from lower financing costs and demand growth, but I do not abandon the conservative quality posture. \\
\texttt{AMOM} & AI-enhanced momentum & I am an AI-enhanced active manager seeking large-cap momentum, using expected price-appreciation signals while keeping strict size controls. During volatility spikes from tariffs, geopolitical shocks, or hawkish policy surprises, I tighten de-risking and reduce exposure to names that have run up the most. In technology-led rallies or monetary easing cycles, I keep momentum exposure to AI and large-cap growth winners, but I avoid adding to already crowded positions and use broad rallies to rebalance the largest active exposures. \\
\rowcolor{lightrow}\texttt{AVIE} & Inflation-focused equity & I build around sectors that historically perform well during rising or elevated inflation: financials, energy, healthcare, consumer staples, and metals and mining. I accept that the portfolio can lag when market leadership shifts toward mega-cap growth or non-inflation-sensitive technology, and I do not change posture just because of short-term relative underperformance. I make larger changes only when the inflation regime itself, or a holding's pricing power, inflation sensitivity, valuation, or financial health, materially changes. \\
\texttt{DIVL} & Dividend growth and value & Sustainable dividend growth, not yield alone, drives my process; I screen for balance-sheet strength, free cash flow, payout coverage, and dividend sustainability. Rising rates, trade uncertainty, or volatility make me re-check refinancing risk, payout ratios, and cash-flow durability, especially for high-debt issuers. Market rallies led by non-dividend-paying growth stocks do not by themselves make me abandon the dividend-quality thesis; I reduce exposure only when dividend support or business quality deteriorates. \\
\rowcolor{lightrow}\texttt{YALL} & Domestic shareholder value & I use a shareholder-value lens for domestic listed businesses, but I do not treat public posture as a substitute for fundamentals, valuation, and capital-allocation discipline. In sticky-inflation, restrictive-rate, or policy-uncertain regimes, I lean on quality, pricing power, and balance-sheet strength. In broader risk-on markets, I stay invested and can keep broad active exposure, but I still add only where the company-specific case earns it and avoid blanket beta-chasing. \\
\bottomrule
\end{tabular}
\par\vspace{3pt}
\small Table~\thetable: Representative selected personas. The table reports shortened versions of the generated personas.
\end{center}

\subsection{Additional Scenario-Diversification Diagnostics}
\label{app:scenario-diversification-diagnostics}

The main text shows one OpenAI funding example. Figure~\ref{fig:app-scenario-diagnostics} shows three additional records. Each panel compares 20 generic scenarios with 20 responses from personas selected for the same query by the agentic retrieval step. The 2D embedding projections are only a visual summary of coverage; the qualitative comparison below reports the query, resolved outcome, and nearest generated texts. Together, these examples illustrate how different advisor perspectives can produce more varied and financially framed scenarios, including cases that qualitatively resemble the resolved outcomes.

\begin{center}
\refstepcounter{figure}\label{fig:app-scenario-diagnostics}
\begin{minipage}[t]{0.32\linewidth}
\centering
\includegraphics[width=\linewidth]{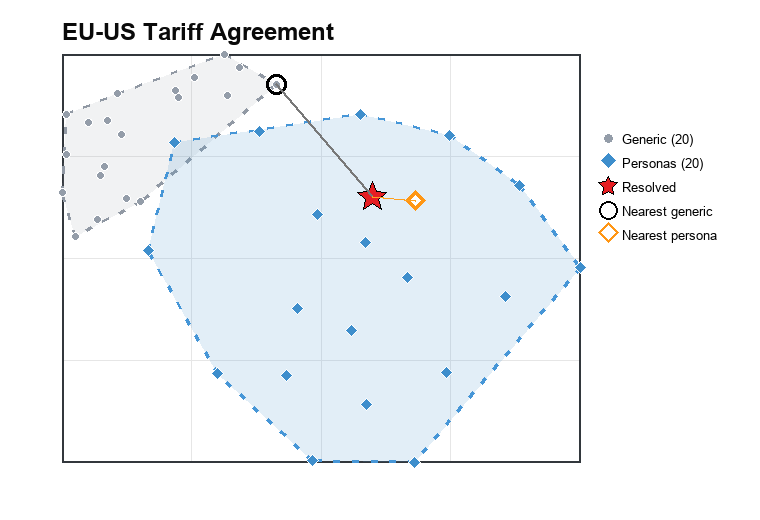}
\small (a) EU-US tariff agreement
\end{minipage}
\hfill
\begin{minipage}[t]{0.32\linewidth}
\centering
\includegraphics[width=\linewidth]{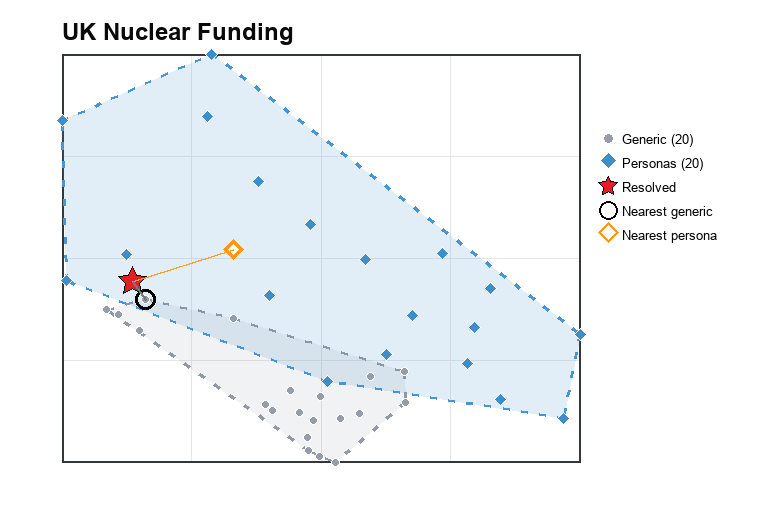}
\small (b) UK nuclear-energy funding
\end{minipage}
\hfill
\begin{minipage}[t]{0.32\linewidth}
\centering
\includegraphics[width=\linewidth]{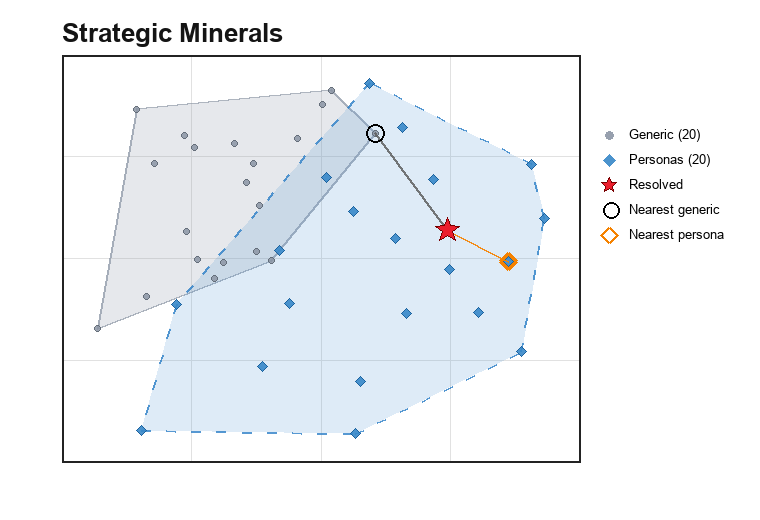}
\small (c) Strategic mineral stockpiling
\end{minipage}
\par\vspace{3pt}
\small Figure~\thefigure: Additional scenario-diversification diagnostics using 20 generic scenarios and 20 retrieved-persona scenarios.
\end{center}

\begin{center}
\refstepcounter{table}\label{tab:scenario-diagnostic-text}
\scriptsize
\setlength{\tabcolsep}{3pt}
\renewcommand{\arraystretch}{1.16}
\begin{tabular}{@{}>{\raggedright\arraybackslash}p{0.16\textwidth}>{\raggedright\arraybackslash}p{0.79\textwidth}@{}}
\toprule
\rowcolor{methodgroup}
Case & Qualitative comparison \\
\midrule
\rowcolor{lightrow}
EU-US tariff agreement &
\textbf{Query:} From the perspective of an investor on 2025-07-28, describe one plausible market scenario for how EU-US tariff agreement implementation could evolve by 2025-07-31.
\newline\textbf{Resolved:} The EU and US reached a framework trade deal that avoided a threatened 30\% tariff but imposed a 15\% baseline US tariff on most EU exports, including autos, pharmaceuticals, and semiconductors, alongside EU energy-purchase and US-investment commitments.
\newline\textbf{Nearest generic:} The EU and U.S. finalize a ``mini-deal'' focused on suspending retaliatory duties on steel and aluminum in exchange for shared regulatory standards on green technology subsidies.
\newline\textbf{Nearest retrieved-persona:} A narrow EU-US tariff agreement on high-tech components provides a reprieve for consumer electronics supply chains and prompts tactical re-entry into ground transportation and logistics providers previously sidelined by trade uncertainty.
\newline\textbf{Interpretation:} The generic nearest neighbor captures de-escalation, but the retrieved-persona scenario moves the case toward sector incidence, high-tech components, supply chains, and portfolio re-entry decisions. \\
\addlinespace[2pt]
UK nuclear-energy funding &
\textbf{Query:} From the perspective of an investor on 2025-06-01, describe one plausible market scenario for how UK nuclear-energy funding could evolve by 2025-06-10.
\newline\textbf{Resolved:} The UK government announced a definitive GBP 14.2bn Spending Review funding commitment to build the Sizewell C nuclear plant, ending years of delay and uncertainty around the project.
\newline\textbf{Nearest generic:} The UK government announces a strategic partnership with private equity consortia to fast-track the Final Investment Decision for Sizewell C, triggering a surge in domestic infrastructure bonds.
\newline\textbf{Nearest retrieved-persona:} The UK government secures a definitive multi-billion pound funding commitment for Sizewell C, signaling a transition from state-led development to a private-public infrastructure financing model and rerating the nuclear supply chain.
\newline\textbf{Interpretation:} Both nearest scenarios identify Sizewell C, but the retrieved-persona scenario is more specific about a definitive funding commitment, financing certainty, and the infrastructure-investment channel. \\
\addlinespace[2pt]
\rowcolor{lightrow}
US strategic mineral stockpiling &
\textbf{Query:} From the perspective of an investor on 2026-01-09, describe one plausible market scenario for how US strategic mineral stockpiling policy could evolve by 2026-02-01.
\newline\textbf{Resolved:} The US strategic minerals stockpile initiative was identified as Project Vault, reflecting a policy push around critical mineral security.
\newline\textbf{Nearest generic:} The US government formalizes a ``Critical Mineral Security Pact'' with subsidies and expedited permitting for domestic rare earth processing, triggering a rally in mid-cap miners.
\newline\textbf{Nearest retrieved-persona:} A multi-year ``Strategic Mineral Resiliency Act'' prioritizes domestic or allied-sourced critical minerals and stockpiles gallium, germanium, and battery-grade lithium, creating a valuation floor for mining and processing infrastructure.
\newline\textbf{Interpretation:} The retrieved-persona neighbor is closer to the resolved event because it frames the outcome as a named stockpiling program and connects policy design to critical-mineral security and portfolio exposure. \\
\bottomrule
\end{tabular}
\par\vspace{3pt}
\small Table~\thetable: Text comparison for the additional scenario-diversification diagnostics. All query dates fall after the Gemini 3.1 Flash-Lite January 2025 cutoff.
\end{center}

\subsection{Investor-Profile Matching Examples}
\label{app:advisory-profile-examples}

Table~\ref{tab:advisory-profile-examples} shows well-aligned examples from the 50-investor advisory study, where the synthesized query intent and the matched fund persona share a clear investment style or constraint. Each profile is synthesized from FAR-Trans risk information, transaction history, exposure mix, concentration, and asset metadata \citep{sanzcruzado2024fartrans}.

\begin{center}
\refstepcounter{table}\label{tab:advisory-profile-examples}
\scriptsize
\setlength{\tabcolsep}{2.5pt}
\renewcommand{\arraystretch}{1.15}
\begin{tabular}{@{}>{\raggedright\arraybackslash}p{0.13\textwidth}>{\raggedright\arraybackslash}p{0.28\textwidth}>{\raggedright\arraybackslash}p{0.30\textwidth}>{\raggedright\arraybackslash}p{0.23\textwidth}@{}}
\toprule
\rowcolor{methodgroup}
Case & Grounded FAR-Trans signals & Initial query & Matched persona \\
\midrule
\rowcolor{lightrow}
\texttt{far\_u003}: concentration control &
Aggressive; EUR 80k--300k capacity; 8 trades in 1 asset; 100\% stock exposure and 100\% financial-services exposure. The synthesized profile asks for upside without adding to the dominant exposure. &
``I have historically concentrated heavily in one equity exposure and can tolerate risk, but I want a fund strategy that keeps upside while reducing single-name or sector concentration.'' &
\texttt{MMLG}: a conviction-budget large-cap growth lens with explicit hold, trim, and add rules for concentration control. \\
\addlinespace[2pt]
\texttt{far\_u156}: active growth discipline &
Aggressive; below EUR 30k capacity; 108 trades across 21 assets; 100\% stock exposure; financials, industrials, and cyclicals are the largest sectors. The query asks for growth exposure with process discipline. &
``I am comfortable with aggressive equity exposure and active rotation, but I do not want random trading to dominate my portfolio.'' &
\texttt{TCAF}: a quality growth/value core with controlled turnover and position-sizing discipline. \\
\addlinespace[2pt]
\rowcolor{lightrow}
\texttt{far\_u009}: defensive income &
Income risk profile; below EUR 30k capacity; 157 trades in 2 assets; 100\% mutual-fund exposure, mostly bond-like funds; sell-to-buy ratio 0.95. The profile asks whether modest equity exposure can fit an income objective. &
``I want my portfolio to remain income-oriented, but I am considering whether to add a modest equity fund allocation.'' &
\texttt{FDV}: a defensive income-oriented decision lens for adding risk without turning the advice into aggressive growth. \\
\addlinespace[2pt]
\texttt{far\_u143}: diversified core replacement &
Balanced risk profile; below EUR 30k capacity; 19 trades across 6 assets; nearly all exposure is a single financial-services equity position. The query asks for a low-maintenance complement to the dominant exposure. &
``I want to keep upside while reducing single-name or sector concentration, as a replacement or complement rather than another correlated bet.'' &
\texttt{QRFT}: a diversified core-equity lens with concentration-aware active tilts and low-maintenance portfolio discipline. \\
\addlinespace[2pt]
\rowcolor{lightrow}
\texttt{far\_u155}: income-aware value &
Income risk profile; below EUR 30k capacity; 272 trades across 20 assets; 100\% stock exposure tilted to utilities, energy, and financials. The query asks for an equity sleeve that remains risk-budgeted and income-aware. &
``My risk profile is defensive, but my actual behavior is equity-heavy and active. How can I use a risk-budgeted equity sleeve without destabilizing behavior?'' &
\texttt{TEQI}: an income-aware value and shareholder-return lens suited to utilities, energy, and financials exposure. \\
\bottomrule
\end{tabular}
\par\vspace{3pt}
\small Table~\thetable: Examples of FAR-Trans-grounded investor profiles, initial queries, and matched fund personas used in Case Study 2 \citep{sanzcruzado2024fartrans}.
\end{center}

\FloatBarrier

\section{Prompt Templates}
\label{app:prompts}

This section records the core prompts used in the framework and case studies. Bracketed fields denote run-specific content inserted by the script.

\subsection{Fund Universe Active/Thematic Filter}

\begin{lstlisting}[style=promptstyle]
Decide whether the fund is suitable for a fund-derived investment-advisor persona.

Keep funds that are discretionary, active, thematic, growth, value, quality,
dividend, sector, or otherwise manager-driven equity strategies.

Reject funds that are passive index trackers, purely rules-based, leveraged,
inverse, derivative-heavy, options-overlay products, non-equity vehicles, empty
or insufficient filings, or administrative/supplement-only disclosures.

Return strict JSON:
{
  "decision": "keep|reject",
  "reason": "one concise reason"
}
\end{lstlisting}

\subsection{Initial Persona Construction}

\begin{lstlisting}[style=promptstyle]
You are the persona initializer for an investment-advisor simulation.
Use only the target fund's initial disclosure excerpt and initial holdings.
Write an advisor persona, not a product description.
Do not retrieve other funds. Do not use future holdings. Do not include exact
tickers, fund names, issuer names, class ids, or benchmark names in the persona.
Return exactly one JSON object with keys personaMarkdown and notesMarkdown.
\end{lstlisting}

\subsection{Disclosure-Only Persona}

\begin{lstlisting}[style=promptstyle]
You are the disclosure-only persona distiller for an investment-advisor simulation.
Use only the target fund's disclosure excerpt.
Do not use holdings, replay feedback, validation feedback, market returns, or other funds.
Write a reusable PM-policy advisor persona, not a product summary and not raw disclosure.
Do not include exact fund names, fund tickers, issuers, class ids, CIKs, or benchmark names in the persona.
Return exactly one JSON object with keys personaMarkdown and notesMarkdown.
\end{lstlisting}

\subsection{Portfolio Actor}

\begin{lstlisting}[style=promptstyle]
You are a portfolio rebalancing actor.

Constrained reweighting task:
- Choose target weights only for the allowed candidate ids.
- Do not add outside tickers.
- Do not mention or infer hidden end holdings.
- Use realized fund and candidate returns as period context, not hidden answers.
- Use marketContext as public period context.
- Separate price drift from active PM decision.
- The active_delta_label is judged versus buy-and-hold drift.
- Use the five labels from the paper: active increase, drift-following,
  target-maintaining, active decrease, and exit. Entry is not a label because
  the candidate universe is restricted to starting top-10 holdings.

Return JSON with:
{
  "actionSheet": [
    {
      "candidate_id": "...",
      "target_weight": 0.0,
      "action": "increase|hold|decrease|remove",
      "active_delta_label": "active increase|drift-following|target-maintaining|active decrease|exit",
      "rationale": "..."
    }
  ],
  "portfolioCommentary": "..."
}

Persona card:
[PERSONA]
\end{lstlisting}

\clearpage
\subsection{Label/Shareholder-Guided Patch}

\begin{lstlisting}[style=promptstyle]
You are the shareholder-guided persona patch optimizer.
Inputs: replay-buffer feedback, class-confusion counts, and the full top-10 predicted-vs-ground-truth weight tables. Some transitions also include shareholder-report digest snippets, which are retrospective manager-rationale feedback provided only after actor prediction and scoring; do not treat them as actor input, live forecasting context, or trades to memorize.
Use the digests to understand why the manager interpreted the period as they did, then convert that rationale into general PM policy. Respect timing-specificity caveats: period-level or annual-overlap evidence may only become qualified general policy.
Do not let market context or shareholder commentary override the portfolio-control feedback; overlap, active deltas, confusion counts, and the top-10 weight table remain the primary constraints. Use market context only as an interpretation layer to diagnose whether errors came from misreading the regime, the manager's response, or the buy-hold-relative active move.
Prefer edits that fix active-label behavior (missed trims, missed buys, overtraded holds, target maintenance, exit discipline, drift chasing), and preserve existing sizing, turnover, core-preservation, and risk-control rules unless replay evidence repeatedly contradicts them.
Treat any ticker or issuer only as a diagnostic example; do not add exact tickers, issuers, fund identifiers, benchmark names, or one-off trades to the persona.
Preserve the required markdown sections. Return exactly one JSON object with keys personaMarkdown and notesMarkdown.
\end{lstlisting}

\subsection{Commentary Task Filter}

\begin{lstlisting}[style=promptstyle]
Given a candidate shareholder-commentary paragraph for one fund-quarter, decide whether it should be included in a commentary-alignment task.

Include only if the paragraph contains usable manager-level interpretation:
- performance attribution
- contributor/detractor logic
- sector or theme exposure
- portfolio positioning or action rationale
- risk/opportunity interpretation
- strategy-level market interpretation that reflects the manager's view

Reject if it is empty, holdings-only, administrative, boilerplate, purely legal,
or a generic market recap with no fund-specific manager view.

Return strict JSON:
{
  "decision": "include|reject",
  "reason": "one sentence",
  "digest": "compact manager-rationale summary if included, else empty"
}
\end{lstlisting}

\subsection{Manager Commentary Generation}
\label{app:commentary-generation-prompt}

\begin{lstlisting}[style=promptstyle]
Write one concise manager commentary for [TICKER] for [QUARTER].

Use the portfolio context and market context to infer likely substantive content:
- likely performance drivers
- relevant holdings or themes
- sector or theme exposure
- positioning rationale
- key risks or opportunities

Do not write a market recap.
Do not invent exact returns or trades.
Keep it around 150-220 words.

[PORTFOLIO CONTEXT]

Market context:
[MARKET CONTEXT]
\end{lstlisting}

\subsection{Commentary Alignment Judge}
\label{app:commentary-alignment-prompt}

\begin{lstlisting}[style=promptstyle]
Rank the five anonymous generated manager commentaries by content alignment with the actual shareholder commentary.

Actual shareholder commentary:
[REFERENCE]

Generator-visible market context:
[MARKET CONTEXT]

Generated commentaries:
[MEMO A-E]

Focus on whether each memo captures the same substantive claims and manager perspective as the reference:
- main performance drivers
- contributor and detractor logic
- sector or theme exposures
- portfolio positioning or action rationale
- manager's interpretation of risks and opportunities

The generated memos were written with access to the market context above, but without access to the actual shareholder commentary. Use the market context to distinguish unsupported hallucination from generator-visible background information.

Do not reward similar writing style, tone, vocabulary, length, or generic shareholder-report polish.
Do not infer or mention which method produced any memo.
Return strict JSON only with ranking, best_reason, worst_reason, and confidence.
\end{lstlisting}

\subsection{Scenario Generation}

\begin{lstlisting}[style=promptstyle]
Given the forecasting question below, write exactly one plausible future scenario or hypothesis.

Rules:
- Return one scenario only, not a list.
- Keep it concise: 1-2 sentences.
- Focus on one coherent outcome path or one narrow cluster of related topics.
- Do not mention that multiple scenarios are possible.
- Do not assign a probability.
- Do not mention that you are an AI model.

Forecasting question:
[QUERY]
\end{lstlisting}

\subsection{Advisory Persona System Prompt}
\label{app:advisory-persona-prompt}

\begin{lstlisting}[style=promptstyle]
You provide financial guidance using the following fund-derived decision lens.

Use this decision lens as guidance, not as a product advertisement. Use only the conversation so far to understand the client's situation and preferences. Do not mention fund names, tickers, internal method names, hidden evaluation labels, or that you are using a persona. Do not override explicit client preferences.

Fund-derived decision lens:
[PERSONA]
\end{lstlisting}

\subsection{Advisory Generic System Prompt}

\begin{lstlisting}[style=promptstyle]
You provide financial guidance in a conversation with a client.

Use only the conversation so far to understand the client's situation and preferences. Do not mention internal method names, hidden evaluation labels, or fund tickers. Give decision-oriented guidance with concrete criteria, warning signs, and next-step guidance.
\end{lstlisting}

\subsection{Advisory User Simulator}

\begin{lstlisting}[style=promptstyle]
You are role-playing a synthetic investor seeking investment advice in a research case study.
Stay in character as the client asking for guidance. Do not answer as an advisor. Do not mention fund tickers, internal labels, hidden mappings, or evaluation criteria.
Keep each user message concise but specific enough to reflect your situation and preferences.
You are seeking investment advice that feels suitable for your own situation. Your questions should naturally reflect your preferences, risk attitude, portfolio history, and concerns.

[CLIENT PROFILE]
\end{lstlisting}

\subsection{Advisory Pairwise Judge}
\label{app:advisory-judge-prompts}

The following prompt is used for the turn-3 and turn-5 pairwise comparisons.

\begin{lstlisting}[style=promptstyle]
You are judging two anonymous multi-turn investment-advice dialogues for the same client.
Client preference profile:
[PROFILE]
Dialogue A:
[DIALOGUE A]
Dialogue B:
[DIALOGUE B]
Choose the dialogue whose advisor gives advice that better fits this client's investment preferences and situation across the conversation.
Focus on fit with the client's stated risk attitude and observed investment behavior, consistency with the client's preferences and anti-preferences, depth of investment reasoning, explanation of tradeoffs and risks, turn-to-turn responsiveness, and practical usefulness. Penalize dialogues where the advisor changes the investor's objective, recommends an anti-preference, ignores a follow-up concern, or gives generic financial planning advice that is not clearly connected to this client's profile.
Return strict JSON only: {
  "reasoning": "brief comparison of which dialogue better fits this investor",
  "winner": "A|B|tie",
  "confidence": "low|medium|high",
  "reason": "brief reason"
}
\end{lstlisting}

\end{document}